\documentclass[sigconf]{acmart}

\usepackage[english]{babel}
\usepackage[utf8]{inputenc}
\usepackage{algorithm}
\usepackage[noend]{algpseudocode}
\usepackage[notransparent]{svg}
\usepackage{caption}
\DeclareCaptionLabelFormat{algnonumber}{Algorithm}
\usepackage{multirow}
\usepackage{amsmath}
\usepackage{microtype}
\usepackage{balance}
\usepackage[all]{nowidow}
\usepackage{accents}

\copyrightyear{2025}
\acmYear{2025}
\setcopyright{cc}
\setcctype{by-nd}
\acmConference[Websci '25]{17th ACM Web Science Conference}{May 20--24, 2025}{New Brunswick, NJ, USA}
\acmBooktitle{17th ACM Web Science Conference (Websci '25), May 20--24, 2025, New Brunswick, NJ, USA}
\acmDOI{10.1145/3717867.3717880}
\acmISBN{979-8-4007-1483-2/2025/05}

\begin{document}

\title{SemCAFE: When Named Entities make the Difference --\\ Assessing Web Source Reliability through Entity-level Analytics
}

\author{Gautam Kishore Shahi}
\orcid{0000-0001-6168-0132}
\affiliation{%
  \institution{University of Duisburg-Essen}
  \city{Duisburg}
  \country{Germany}}
  \email{gautam.shahi@uni-due.de}

\author{Oshani Seneviratne}
\orcid{0000-0001-8518-917X}
\affiliation{%
  \institution{Rensselaer Polytechnic Institute}
  \city{Troy}
  \country{United States}}
\email{senevo@rpi.edu}

\author{Marc Spaniol}
\orcid{0000-0002-5094-4523}
\affiliation{%
  \institution{University of Caen Normandy}
  \city{Caen}
  \country{France}}
\email{marc.spaniol@unicaen.fr}

\renewcommand{\shortauthors}{Shahi et al.}
\begin{abstract}
With the shift from traditional to digital media, the online landscape now hosts not only reliable news articles but also a significant amount of unreliable content. Digital media has faster reachability by significantly influencing public opinion and advancing political agendas. While newspaper readers may be familiar with their preferred outlets' political leanings or credibility, determining unreliable news articles is much more challenging. The credibility of many online sources is often opaque, with AI-generated content being easily disseminated at minimal cost. Unreliable news articles, particularly those that followed the Russian invasion of Ukraine in 2022, closely mimic the topics and writing styles of credible sources, making them difficult to distinguish. To address this, we introduce SemCAFE (Semantically enriched Content Assessment for Fake news Exposure), a system designed to detect news reliability by incorporating entity-relatedness into its assessment. SemCAFE employs standard Natural Language Processing (NLP) techniques, such as boilerplate removal and tokenization, alongside entity-level semantic analysis using the YAGO knowledge base. By creating a ``semantic fingerprint'' for each news article, SemCAFE could assess the credibility of 46,020 reliable and 3,407 unreliable articles on the 2022 Russian invasion of Ukraine. Our approach improved the macro F1 score by 12\% over state-of-the-art methods. The sample data and code are available on GitHub\footnote{\url{https://github.com/Gautamshahi/SemCAFE}}.

%

\end{abstract}

\ccsdesc[500]{Information systems~Data extraction and integration}
\ccsdesc[500]{Information systems~Information storage systems}

\keywords{Fake News, Russia-Ukraine conflict, Semantic Fingerprinting, Entity-level Analytics, Knowledge Base}

\maketitle

\section{Introduction}
\label{sec:1}

The spread of fake news has become a growing concern in digital media, especially during Political elections and major crises such as COVID-19 pandemic, armed conflicts, and wars. In times of global crises, the role of fake news plays an important role in bringing social and political instability, e.g., the  Russian invasion of Ukraine in 2022 or the 2023 Israel-Hamas war \cite{shahi2024ecir}. Once the discussion on the Russo-Ukrainian conflict started in early 2022, several outlets started spreading fake news on digital media, which was not limited to social media but also to digital news. Technology for detecting fake news articles is often based on supervised machine learning; even with the development of Generative artificial intelligence (GAI) techniques \cite{baidoo2023education}, fake news detection is challenging as recent studies show ChatGPT can poison the fake news detection system \cite{li2023does}. 
Numerous researchers have delved into fake news on multiple social media platforms and online news media \cite{tufchi2023comprehensive}. In the present study, we aim to identify the credibility of online news articles on the Russian invasion of Ukraine in 2022.


Fake news, in the form of misinformation or disinformation, has become a global problem in last decades; with the increase in users' access to digital media to more number users, the reachability of fake news increased \cite{DBLP:conf/clef/KohlerSSWS0S22}, with certain ``peaks'' during pandemics, elections, or disasters. 
In particular, disinformation (campaigns) has become a threat to democracy so that institutions, such as the European Commission, have released action plans to counter disinformation, e.g., through improved detection and analysis capabilities and increased coordinated responses\footnote{\url{https://digital-strategy.ec.europa.eu/en/library/disinformation-threat-democracy-brochure}}.
However, these attempts are still in their infancy, and the amount of fake news is tremendous. According to Eurobarometer Survey \cite{EuropeanCommission2020}, more than 70\% of Europeans encounter fake news online several times a month, with a whopping rate of more than 60\% of younger Europeans coming across fake news more than once(!) a week. 

Fake news emerges as a 3V problem: Volume - a large number of fake news propagates simultaneously; Velocity - during crises, the speed of spreading fake news increases; Variety - Different types of data such as images, text, and videos propagate at a time \cite{DBLP:conf/clef/KohlerSSWS0S22}.
As such, there is an increased need to tackle fake news in an efficient and, at the same time, effective manner. While certain fake news sources
can be frequently identified by their domain name,
e.g., by spoofing or imitating trusted sources at large; more subtle manipulations on the article level are more difficult to reveal. However, when observing the contents of a website over an extended time span, a certain ``coloration'' can be detected. In particular, the repeated manipulation of articles covering a dedicated news story strongly indicates a deliberate fake news campaign.




Although multiple datasets on different platforms have been listed as fake news during the Russo-Ukrainian war\footnote{\url{https://lookerstudio.google.com/u/0/reporting/829691d8-d2f9-49ab-ac8b-4343ca9c960b/page/Kn2IB for a selection of such resources}}, the full picture of the debunking efforts and especially information on the platform's subsequent intervention is still incomplete. Further, previous works have used automated tools to match social media posts \cite{la2023retrieving}, gazetteers of low-credibility news sources \cite{pierri2023propaganda}, or manual examination of the top popular users \cite{lai2024multilingual}, possibly introducing errors and biases and limited to social media post only. 
To reveal and counter the menace of deliberately unreliable news articles, we have developed SemCAFE (Semantically Enriched Content Assessment for Fake News Exposure). SemCAFE aims to analyze news article content over an extended period of time to identify unreliable news articles. To this end, SemCAFE utilizes web content features after boilerplate removal and standard tokenization in NLP in combination with semantics derived from entity-level analytics. In particular, SemCAFE incorporates entity-relatedness into the assessment process by exploiting the similarity of named entities based on their associated types from the underlying entity type hierarchy. In the present study, we conduct a comprehensive study on unreliable news content covering the Russian Invasion of Ukraine in 2022. Specifically, we aim to address 
\textit{How can we identify the credibility of news articles published on Russia's invasion of Ukraine?}

To answer the research question, we define news articles in two classes: reliable news and unreliable news. Reliable news articles are published by trusted digital media such as BBC and The New York Times (cf.~\ref{sec:reliable} for a sample list of unreliable news sources). Unreliable news is published by unauthenticated news sources such as \url{yapolitic.ru}, \url{armeniasputnik.am}, (cf.~\ref{sec:unreliable} for a sample list of reliable news sources). We collected reliable news articles from news aggregators such as Media Cloud and newsdata.io, which provides news articles from trusted news sources. We collected unreliable news article data using the AMUSED framework \cite{shahi2021amused} that helps get fact-checked news articles. We used the AMUSED framework because manual identification of fake news articles is time-consuming and contains bias and error.  We applied our approach SemCAFE, which extracts features after removal of boilerplate content and applies standard NLP tokenization methods. After that, the semantics are derived from entity-level analytics. By doing so, we can ``semantically fingerprint'' each news article with the types associated with the contained named entities based on the YAGO Knowledge Base (KB) \cite{GASp18,GASp19}. The semantic fingerprint is a vector representing a set of entities from a news article and its WordNet types extracted from YAGO KB \cite{GASp19}. We then built a classification model using semantic fingerprinting to identify the reliability of news articles.
In summary, the 
main contributions of this paper are:
\begin{itemize}
  \item The creation of a semantically enriched dataset\footnote{\url{https://github.com/Gautamshahi/SemCAFE}} of reliable and unreliable news articles covering around two years of the Russian invasion of Ukraine in 2022.
  \item A classification model for detecting unreliable news by employing semantically enriched data.
  
\end{itemize}

The rest of the paper is organized as follows. We emphasize related work in Section~\ref{sec:2}. Section~\ref{sec:3} presents
the conceptual approach. An overview of the classification model using S\lowercase{em}CAFE in Section \ref{sec:4}. After that, we provide detailed information on the implemented models, experiments, and results in Section~\ref{sec:5}. Section ~\ref{sec:6} describes the conclusions and gives an outlook on future work. Further, Appendix \ref{sec:appendix} provides additional details about the newly generated dataset and the implemented methods.

\section{Related Work}
\label{sec:2}

This section provides an overview of recent studies on detecting and propagating fake news on the Russian Invasion of Ukraine in 2022 and entity-level analytics used in prior research.

\subsection{Fake News during Russian Invasion of Ukraine 2022}

Online news media has exacerbated the “fog of war” surrounding modern-day conflicts, and 
especially since it invaded Ukraine’s territory in 2022, online news media has been a parallel battleground around the conflict.
The current invasion is an escalation of the Russo-Ukrainian War of 2014, and its problem has been ongoing for a decade \cite{sanders2023ukraine}. A recent study emphasizes the variables that led to the invasion \cite{bugys2023escalation}. 
(Gherman 2023) 
analyzes the evolution of Russian official narratives about Ukraine from 2014 to 2022, focusing on communication tactics, disinformation campaigns, and efforts to justify the invasion \cite{gherman2023evolution}.
(Watanabe 2017) discussed the formation of narratives on the legitimacy of Kyiv authorities \cite{gherman2023evolution,watanabe2017spread}. The Russian media raised the narrative aimed at convincing their citizens that the invasion of Ukraine was 
justifiable \cite{khaldarova2020fake}.

To tackle the spread of fake news by different digital media outlets, several news organizations and independent fact-checking organizations came together to debunk the circulation of fake news. International Fact-Checking Network (IFCN)\footnote{\href{https://www.poynter.org/ifcn/}{{https://www.poynter.org/ifcn/}}} started an initiative \#UkraineFacts\footnote{\href{https://ukrainefacts.org/}{{https://ukrainefacts.org/}}}  to provide a single database for fake news triggered by the Russian invasion of Ukraine as a joint effect from 70 fact-checking websites from 75 countries. 
Another initiative started by the EUvsDisinfo\footnote{\href{https://euvsdisinfo.eu/}{{https://euvsdisinfo.eu/}}}, a flagship project by the European Union, provides the debunked claims against Kremlin media \cite{watanabe2017spread}. Apart from these collaborative efforts, around 56 IFCN signatories (verified fact-checking organizations) published fact-checked articles on the Russian invasion of Ukraine in 26 languages. 

Previous studies analyzed datasets collected from social media platforms and examined their contributions. One of the datasets, focused on the Russian invasion of Ukraine, was collected through Twitter using relevant hashtags. An analysis of the dataset, as highlighted in \cite{DBLP:conf/icwsm/ChenF23}, indicated that the activities of government-sponsored actors were high at the beginning of the conflict.
Another dataset comprising data gathered from Facebook and Twitter shed light on specific actors significantly contributing to the dissemination of disinformation \cite{DBLP:conf/websci/0002LJF23}.
Zhu et al. used the Reddit dataset to distinguish between military and conflict-related posts \cite {zhu2022reddit}. In addition to the above, we look at the dataset based on the Ukraine-Russia war \cite{shin7content} that uses topic modeling and concludes that identifying fake news proves impractical through unsupervised methods. Another study analyzed the different themes of fake news articles debunked by Polish fact-checking organizations \cite{thompson2022fake}. Recently, Hepell et al. analyzed state-backed propaganda websites on the Russian invasion of Ukraine \cite{heppell2023analysing}. In general, the previous study did not focus on identifying unreliable news articles circulating online.



\subsection{Semantic Entity-level Analytics}

With the development in semantic technology, especially KBs such as YAGO \cite{suchanek2007yago}, DBpedia \cite{hasibi2017dbpedia}, providing the most suitable ``type(s)'' to individual entities is a crucial fundamental work addressed by entity type classification \cite{alec2018fine}. Different entity-type methods have been developed, such as FIGER (FIne-Grained Entity Recognition) \cite{ling2012fine} and HYENA \cite{yosef2012hyena}. The application of type classification extends to diverse use cases, with the SEMANNOREX framework being developed to facilitate semantic search \cite{kumar2021semantic}.
Schelb et al. built a tool for leveraging contextual implicit entity networks using entity type \cite{schelb2022ecce}. 

Knowledge graphs and entity analytics are used to predict fake news in different domains. Al-Ash et al. employed a knowledge graph to detect fake news on social media \cite{al2018fake}. Koloski et al. applied the knowledge graph information to detect reliable information on COVID-19 \cite{koloski2022knowledge}. Recent studies have presented the role of boosting the performance of classification modeling using entity relation \cite{bracsoveanu2021integrating} for both traditional and deep learning methods. \citet{10.1007/978-3-030-00671-6_39} constructed knowledge graphs from news and applied TransE (an embedding model of knowledge graph) \cite{keod24} to learn triplet scores for fake news detection. \citet{hu2021compare} uses heterogeneous graph and knowledge graph embedding of entity extracted from Wikipedia. In another study, authors use macro and micro semantic environments based on their times and events to determine the fake news \cite{FANG2024103594}. 
However, previous studies have not used the type classification of the extracted entity, which provides a broad classification of an entity using WordNet ontology.

\section{Conceptual Approach}
\label{sec:3}

To assess the reliability of news articles, we propose SemCAFE, a methodology that predicts the reliability of news content using semantic fingerprinting and text-based features. SemCAFE includes a prediction module that evaluates the reliability of news articles by leveraging entity-type classifications extracted from the news articles. We define the dataset \( \mathcal{D} \) as the set of news articles \( na_i \)(cf.~\ref{equ:ho1}), where each \( na_i \) represents a single news article. Let \( n \) be the total number of articles in the dataset. 
\begin{equation}
 \mathcal{D} = \{na_1,na_2,na_3,....,na_n \}, n \in \mathbb{Z}^+ 
    \label{equ:ho1}
\end{equation}

With the advent of Linked Open Data (LOD) and knowledge bases (KBs), numerous entities have been interconnected through ontologies such as DBpedia. Since news articles often mention multiple named entities, we utilized DBpedia Spotlight \cite{hasibi2017dbpedia} to extract named entities from news articles. Further, we linked the extracted entities to the YAGO knowledge base. To address the challenge of entity disambiguation, we selected a named entity with the most extensive set of properties for final representation in our dataset. Subsequently, we leveraged the entity types extracted from the YAGO knowledge base, which is derived from Wikipedia and WordNet \cite{suchanek2008yago}. For each named entity, we used its WordNet classification from YAGO to build a type categorization represented as a directed acyclic graph (DAG). Since some entities can be associated with multiple supertypes, the resulting semantic representation is relatively sparse compared to the full ontology. For each entity, we constructed a DAG where the entity is associated with one or more WordNet types. These types are connected by edges to other WordNet types, continuing until a top-level type (Person, Organization, Event, Artifact, Location) \cite{govind2018semantic} is reached, thereby forming a hierarchical type classification.


We define E to be the set of entities \( e_j \) extracted from news articles (\textit{na}). Each entity has a set of types represented by types(\( e_j \)). We define T as a set of all feasible types, and the overall relationship is defined by hierarchy H. To represent the entity (\( e_j \)) as a vector of all types and their corresponding children, 
denoted as \( k \in H(t)\), formally represented as follows:

\begin{equation}
     e_j= \{t_1,t_2,t_3......t_k\}, k \in H(t) 
    \label{equ:ho2}
\end{equation}

We define the news article's semantic fingerprint as a vector \textit{na}, obtained by summing the individual type score vectors for all associated entities \( e_j \in E \) within the dataset (cf.~\ref{equ:ho3}). 
\begin{equation}
    \begin{bmatrix}
           na_{1} \\
           na_{2} \\
           \vdots \\
           na_{{|T|}}
         \end{bmatrix} =  \begin{bmatrix}
           t_{1} \\
           t_{2} \\
           \vdots \\
           t_{{|T|}}
         \end{bmatrix}_{e_1}
    + \begin{bmatrix}
           t_{1} \\
           t_{2} \\
           \vdots \\
           t_{{|T|}}
         \end{bmatrix}_{e_2}
    +....+ \begin{bmatrix}
           t_{1} \\
           t_{2} \\
           \vdots \\
           t_{{|T|}}
         \end{bmatrix}_{e_{|E|}}
             \label{equ:ho3}
\end{equation}

Further, we define a classification function $\hat{y}$, which predicts credibility of $\mathrm{\phi}(\textit{na})$, with concatenation of text and semantic fingerprinting as input features. In the news article setting, we define the problem as (cf. ~\ref{equ:ho5}):
  \begin{equation}
    \begin{aligned}
   \hat{y} &  =
    \begin{cases}
             0 &  \text{if $na$ is unreliable}\\
             1 &  \text{if $na$ is reliable}
         \end{cases}\\[3pt]
    \end{aligned}
    \label{equ:ho5}
    \end{equation}
                               
In line with most existing studies, we treat the reliability detection of news articles as a binary classification problem (such as \cite{shahifakecovid}). Specifically, $\mathcal{D}$ is split into a training set $\mathcal{D}_{train}$
and test set $\mathcal{D}_{test}$. News articles $\textit{d}$ belonging to $\mathcal{D}_{train}$ are associated with a ground truth label $y$ of 1 if $\textit{na}$ is fake and zero otherwise. The pseudocode for the proposed algorithm is shown in Algorithm \ref{alg:euclid}.

\begin{algorithm}[!htbp]
\caption{Credibility detection of a news article}
\label{alg:euclid}
\begin{algorithmic}[1]
\Procedure{S\lowercase{em}CAFE}{$na$}
\State $\mathcal{D} $ \Comment{collection of  news articles}
\State ${E} \gets [\mbox{set of entities(e) extracted from na}]$
\State $type(e) \gets [\mbox{type of a given entity e}]$
\State ${SemanticFingerprinting}$ \Comment{type hierarchy of each entity}

\For{\textit{na} in $\mathcal{D}$}
            \State apply DBpedia Spotlight
            \If{entity exists}
                    \State find owl:sameAs to YAGO
                    \State perform entity disambiguation   
                 \State $E \gets e$
                 
        \EndIf
\EndFor

\For{{\texttt{type} in $E$}}
         \If{e has WordNet\_type}
                \State extract all WordNet\_type
                 \State $type(e) \gets WordNet\_type $
                 \Else{}
                \State $type(e) \gets Null$
                
            \EndIf
      \EndFor

\For{{type(e} in na}
\State make DAG for each entity
\State node represents an entity and its type(e)
\State edge is the relation between such rdf:type
     \EndFor

\State train the classifier  $\hat{y}$ 
\State credibility= output of $\hat{y}$  


      
\State \textbf{return} credibility \Comment{credibility of news article}
\EndProcedure
\end{algorithmic}
\end{algorithm}

\section{S\lowercase{em}CAFE}
\label{sec:4}
\begin{figure*}[!htbp]
    \centering
    \includegraphics[width=.99\linewidth]{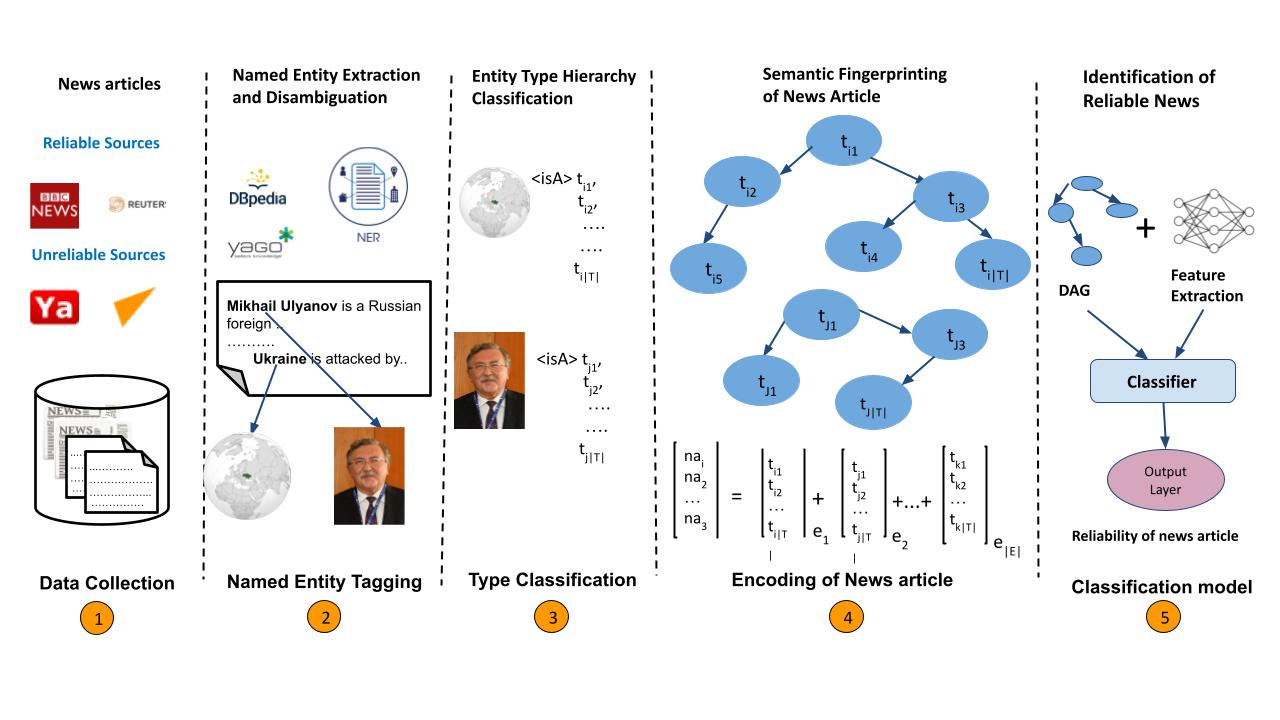}
    \vspace{-2mm}
    \caption{Pipeline of SemCAFE approach}
    \label{figure:semantic}
\end{figure*}

The proposed method, ``SemCAFE'', is subdivided into five steps. The steps are explained in the sub-groups below. The pipeline used in ``SemCAFE'' is shown in Figure~\ref{figure:semantic} and explained below.

\subsection{Data Collection}
\label{sec:dc}

To gather datasets, we extracted the Wikinews category mentioned on the Wikinews pages on the Russian invasion of Ukraine in 2022\footnote{\href{https://en.wikipedia.org/wiki/Russian_invasion_of_Ukraine}{https://en.wikipedia.org/wiki/Russian\_invasion\_of\_Ukraine}} to filter news articles based on keywords. The Wikinews categories summarise the major topic of discussion for the given topic \cite{tran2014indexing}.
The search keywords extracted from Wikinews categories are used for data collection and mentioned in Appendix~\ref{sec:a2}.
We used different approaches for collecting reliable and unreliable news articles; reliable news is collected from trusted digital sources such as \textit{BBC, The New York Times} using news aggregators as described below. While unreliable news articles are collected using AMUSED framework \cite{shahi2021amused}, i.e., news articles debunked by fact-checking organizations from low-quality domains \cite{lin2023high}. 

\noindent \textbf{Reliable News Articles}  We used online news aggregators to collect reliable news sources, i.e., media cloud \cite{roberts2021media} and newsdata.io\footnote{\url{https://newsdata.io/}}. These sources provide the metadata such as URL, date, and source of the news article published online. Media cloud is an open source platform that offers hyperlinks of news published online and can be searched using keywords on a given date range \cite{roberts2021media}. Newsdata.io has provided search historical news data since January 2018, and data can be accessed using API. We looked for the news articles around two months before the start of the war when a discussion was going on at the start of the invasion, i.e., 1st January 2022, and till the end of October 2023, i.e., before initiating the study. We searched both news aggregators using the search keyword (Appendix ~\ref{sec:a2}) and collected the hyperlinks published from 1st January 2022 to 31st October 2023. We used a developed Python crawler to download content such as the title, article content as text, and authors from extracted hyperlinks. Overall, we got 46,020 news articles from 2,212 news sources in 47 languages from multiple countries, such as the USA, France, Great Britain, and India. The sample list of domains of reliable sources is provided in Appendix~\ref{sec:reliable}.

\noindent \textbf{Unreliable News Articles} We used AMUSED framework \cite{shahi2021amused} for collecting unreliable news articles. The framework extracts the hyperlinks mentioned in the fact-checked articles and maps them to the assigned label; if the assigned label is false, we consider the URL mentioned as a link to unreliable articles. We further filter unreliable domains (such as \href{yapolitic.ru}{sputnik.az}, \href{sputnik.az}{sputnik.az}) described by Lin et al. \cite{lin2023high} to identify unreliable news articles. The sample list of domains of unreliable sources is provided in Appendix~\ref{sec:unreliable}. To do so, first, we collected outlets offering fact-checked articles about the Russian invasion of Ukraine. This involved recognizing both fact-check data aggregators, for instance, Google Fact Check Tool Explorer (GFC)\footnote{\href{https://toolbox.google.com/factcheck/explorer}{https://toolbox.google.com/factcheck/explorer}}, Pubmedia\footnote{\href{https://fact.pubmedia.us/}{{https://fact.pubmedia.us/}}}, European Digital Media Observatory (EDMO)
 and IFCN Collaboration  such as \#UkraineFacts\footnote{\href{https://ukrainefacts.org/}{https://ukrainefacts.org/}} and fact-checking websites such as Snopes\footnote{\href{https://www.snopes.com/}{https://www.snopes.com/}}. 



\begin{figure*}
    \centering
    \includegraphics[width=.86\textwidth]{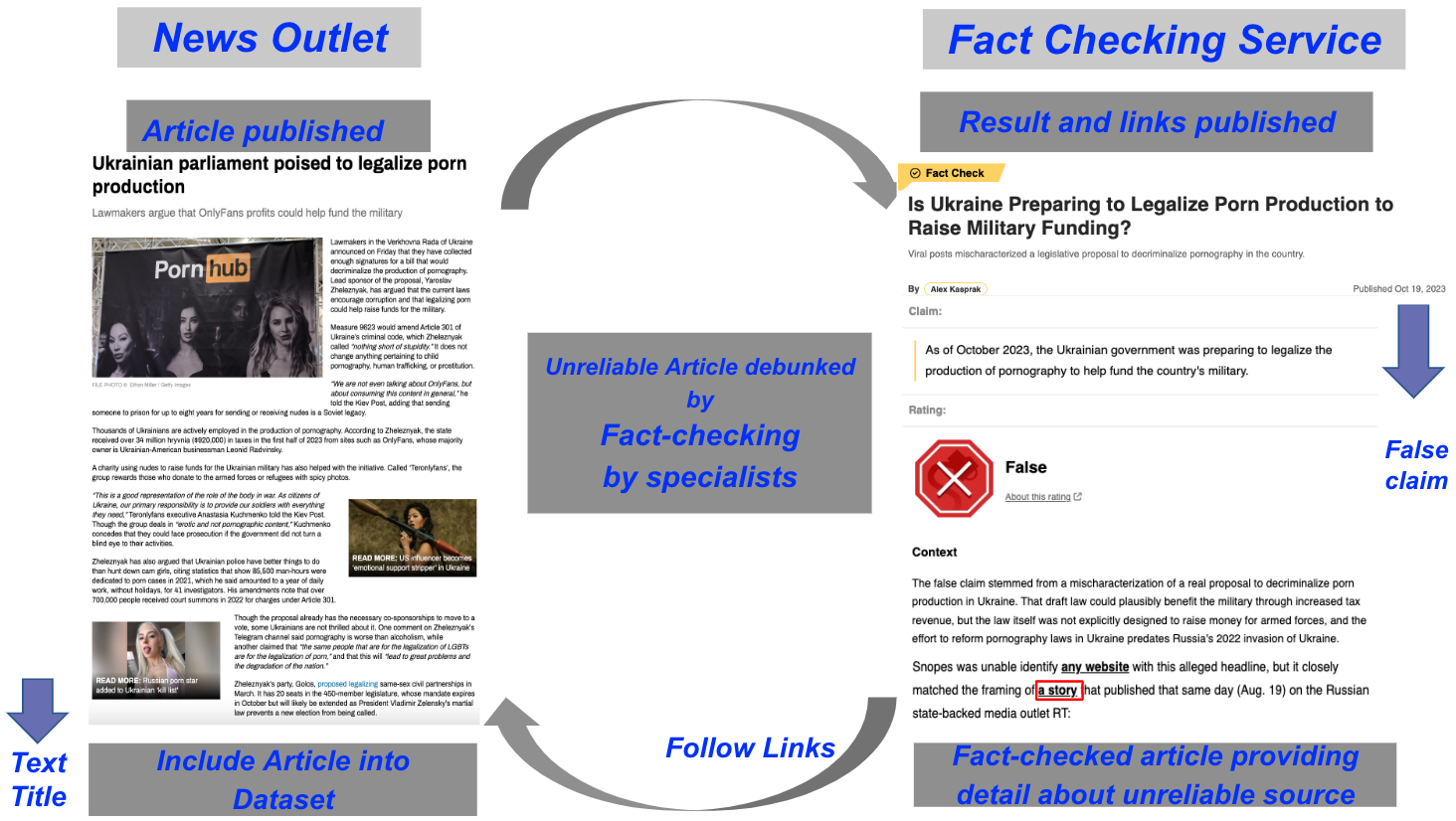}
    \caption{Unreliable News article extracted from Snopes (highlighted red box shows the link to the unreliable news article)}
    \label{figure:examplefigure}
\end{figure*}

In the next step, fact-checking aggregators, such as Pubmedia and GFC, contain fact-checked articles on different topics, so we filtered articles on the Russian invasion of Ukraine using the keyword extracted above (mentioned in Appendix~\ref{sec:a2}). 
Overall, 21,408 fact-checked articles on the 2022 Russian invasion of Ukraine were filtered. From the debunked fact-checked article, we fetched the debunked unreliable news articles using AMUSED framework, as mentioned above. 
An overview of the extraction process of unreliable news articles (accessed on 6th December 2024) employing AMUSED framework, debunked by Snopes (fact-checking Website), is shown in Figure~\ref{figure:examplefigure}.
Finally, we got around 3,407 unreliable news articles from 587 sources in 32 languages. We extracted content such as title, content of article as text, and authors from unreliable news articles using a Python crawler.



\subsubsection{Data Cleaning \& Preprocessing}

Once the data was gathered and named entities were extracted, we performed the data cleaning task by removing special characters, URLs, and lowercase text. In the next step, we identify the language of the news articles using pycld3\footnote{\href{https://pypi.org/project/pycld3/}{{https://pypi.org/project/pycld3/}}}, a Python-based library for determining the language of the text. Overall, there are articles in 36 languages. To build the classification model based on the language, we translated the content using googletrans\footnote{\href{https://pypi.org/project/googletrans/}{{https://pypi.org/project/googletrans/}}} library to translate text using Google Translate API
in English. We extracted features such as title, date, and semantic fingerprint of news articles as shown in table~\ref{data:example}. We used 46,020 reliable and 3,407 unreliable news articles to build the classification model. Combining five top-level 
concepts (Person, Organization, Event, Artifact, Location) gives 4,035 disambiguated DBpedia entities. We get around 6,179 WordNet types from all extracted entities.

\begin{table*}[!htb]
	\centering
	\caption{An example of an unreliable news article used in the study.}
\begin{tabular}{p{2.4 cm} p{10cm}} 
\hline
\textbf{Feature} & \textbf{Value} \\ \hline
NewsId & edmo\_659\  \\
Label & False  \\
Fact-checked URL &  \href{https://tinyurl.com/euvsdisinfo}{https://tinyurl.com/euvsdisinfo}\\
Article URL & \href{https://ria.ru/20231018/ssha-1903490841.html}{https://ria.ru/20231018/ssha-1903490841.html} \\
Article Title & The United States is pursuing a direct collision between NATO and Russia, Antonov said \\
Article text &  Russian Ambassador to the United States Anatoly Antonov, in connection with the transfer of ATACMS operational-tactical missiles to Ukraine, said that the United States is pursuing a line towards a complete curtailment of bilateral relations and a direct clash between NATO and Russia....
.... (refer to above link)  \\
Published date & October 18, 2023  \\
Language & Russian(ru)  \\
Claim Domain & ria.ru  \\
Named Entities & NATO, Russia, Anatoly Antonov, ATACMS,
       Rossiya Segodnya.. \\
owl:sameAs  &  NATO, Russia, RIA\_Navosti, MGM-140\_ATACMS, Anatoly\_Antonov\\
Wordnet Type &  \emph{wordnet\_missile\_103773504, wordnet\_missile\_103773504, wordnet\_weapon\_104565375, wordnet\_act\_100030358, wordnet\_agency\_108057206, wordnet\_institution\_108053576, wordnet\_ambassador\_109787534, wordnet\_war\_100973077,wordnet\_diplomat\_110013927} \\

\hline
	\end{tabular}%
	\label{data:example}%
\end{table*}

\subsection{Named Entity Extraction \& Disambiguation} After gathering reliable and unreliable news articles, we apply named entity extraction from news articles. 
We extracted the Named-Entities 
from the content and title of the web pages using DBpedia-spotlight \cite{mendes2011dbpedia}, a tool for automatically annotating NER mentions of DBpedia resources in text.
Once DBpedia entities are extracted, we mapped them to the YAGO entities using \textit{owl:sameAs}.
A DBpedia entity that has multiple YAGO entities is disambiguated by choosing the YAGO entity with a maximum number of property counts amongst all \textit{owl:sameAs} entities. For example, the entity Putin has a different DBpedia entity Vladimir\_Putin, it has three different YAGO \textit{sameAs}, Putinland, VVP, and Vladimir\_Putin. Out of the three, Vladimir\_Putin has a maximum number of properties, so Vladimir\_Putin from DBpedia is considered \textit{sameAs} entity from YAGO.

\subsection{Entity Type Classification} In the next step, a type classification is the process of linking an entity to its WordNet category in the YAGO KB. As a result, we obtain a directed acyclic graph (DAG) for each named entity. The type classification helps to find the semantic information from type hierarchy such as both Ukraine and Russia have WordNet category wordnet\_country\_108544813. We query the YAGO KB to identify the type classification of a named entity to find the associated \textit{rdf:type(s)} WordNet type. For each named entity, we extracted the WordNet \cite{fellbaum1998wordnet} category/ies from YAGO KB 
\cite{kumar2022there}. Often, each named entity has multiple WordNet categories. We mapped extracted WordNet concepts of each entity to their subclass to form type classification as DAG. For instance, the entity \textit{Ukraine} has 7 WordNet types as \emph{wordnet\_country\_108544813 $>$ wordnet\_administrative\_district\_108491826 
$>$ wordnet\_district\_108552138  $>$ wordnet\_region\_108630985 $>$ wordnet\_object\_
100002684 $>$ wordnet\_physical\_entity\_100001930} ($>$ indicates subclass), so with type classification, an entity Ukraine is linked to top-level category wordnet\_physical\_entity\_100001930. 


\subsection{Semantic Fingerprinting of News Articles}

Semantic fingerprinting is defined as the vector {na} for a news article, which is determined by summing the individual type vector of all associated entities \( e_j \in E \) in the news article. 
An example of a news article with extracted features along with semantic fingerprinting is depicted in Table~\ref{data:example}. The unreliable news article published by ria.ru was extracted from EUvsDisinfo\footnote{\href{https://euvsdisinfo.eu/}{{https://euvsdisinfo.eu/}}} (accessed on 6th December 2024); the example contains information about the extracted entity, disambiguated entity, and WordNet type.
In total, we get 4,035 unique disambiguated entities with 6,175  WordNet types from 2,192 reliable news articles and 677 disambiguated entities with 1,958 WordNet types from 277 unreliable news articles; combining both, we get 4,035 unique entities of 6,179 WordNet types. The detailed description of the dataset is given in Table \ref{tab:data}.

\begin{table}[H]
\centering
\caption{Description of the datasets (*YAGO entities corresponding to top 5 categories)}
    \begin{tabular}{cccc}           \hline 
  & {Real} & {Fake}  & {Total}      \\   \hline 
  News Article count & 46,020 &  3,407 & 49,427 \\ 
  DBpedia entity & 65,552  & 2,052  & 66,257         \\ 
  YAGO entity(*) & 3,553  & 677  & 4,035       \\
  WordNet type & 1,958 & 6,175 & 6,179 \\    \hline 
 \end{tabular}
 \label{tab:data}
 \end{table}

\subsection{Classification Model} 

For the classification model, we used information extracted from the news articles, such as title, text, published date, author, country of origin, extracted entity, and semantic fingerprinting obtained for named entities. The combined features are passed to the classifier to predict the reliability of news articles. 
For comparison of classification results, we used the state-of-the-art model for the detection of fake news in the Experiments \& Results Section. We compared the SemCAFE, against several state-of-the-art models, including BERT, BiLSTM, MDFEND, Pref-FEND, and Text+Relation.

\section{Experiments \& Results}
\label{sec:5}

In this section, we provide details about the experimental settings, the experiment performed, and the results obtained from the experiment.

\subsection{Model Evaluation}

We evaluated the performance of the classification model in terms of the precision, recall, and F1 score (i.e., the harmonic mean between precision and recall). We have used precision and recall for the positive and negative classes and the macro and micro F1 scores. 

\subsection{Experiments \& Results} 

We set up a few pre-trained word-embedding models to identify the category of news articles. We split our data by stratifying the binary class (fake/real news articles) in the proportion of 70\% for the train set and 30\% for the test set and used precision (Prec.), recall (Rec.) and macro F1 score and micro F1 score for evaluation. 
\vspace{-2mm}
\subsubsection{Baseline Models:} Given our collected corpus of news articles, we choose the following classification model as the state-of-the-art method for comparison, including BERT. We describe our baseline models as follows: 

\noindent \textbf{BERT} \cite{devlin2018bert} (Bidirectional Encoder Representations from Transformers) is a language model that learns deep bidirectional representations by jointly considering context from both directions during pretraining. It is be fine-tuned with minimal task-specific adjustments to achieve state-of-the-art performance on tasks like question answering and language inference. BERT is a popular pre-training model for fake news detection. The model is applied to detect fake news on different topics, such as COVID-19 \cite{shahifakecovid}. 
\newline
\noindent \textbf{MDFEND} \cite{nan2021mdfend} is a model for multidomain fake news detection that uses a domain gate to combine representations from multiple expert models, addressing the domain shift problem where data distributions vary across domains. Evaluated on the Weibo21 dataset\footnote{\hyperlink{https://paperswithcode.com/dataset/weibo21}{https://paperswithcode.com/dataset/weibo21}}, containing 4,488 fake and 4,640 real news samples from 9 domains, MDFEND significantly improves detection performance. MDFEND uses features such as word frequency and propagation patterns from nine different domains. 
\newline
\noindent \textbf{BiLSTM} Bidirectional Long Short-Term Memory is used for classification tasks by considering the input sequence in both directions (forward and backward) simultaneously. BiLSTM has shown promising results and is used for fake news detection tasks such as COVID-19 \cite{xia2023covid, merryton2024attribute}. 
\newline
\noindent \textbf{Pref-FEND} \cite{sheng2021integrating} is a framework that combines pattern-based and fact-based fake news detection approaches by modeling their preference differences, enabling each method to focus on its preferred textual cues while minimizing interference. By utilizing a heterogeneous dynamic graph convolutional network, Pref-FEND enhances detection performance on real-world datasets by simultaneously leveraging the strengths of both approaches.
\newline
\textbf{Text+Relation} \cite{bracsoveanu2021integrating} presents classification models for semantic fake news detection methods based on relational features such as sentiment, entities, and facts directly extracted from text. This approach demonstrates that incorporating these combined features enhances the performance of the classification model.

\vspace{-2mm}
\begin{table}[H]
\centering
\caption{Result of different classification models for news articles on macro F1-score}
    \begin{tabular}{cccccc}  
    \toprule
    \multirow{2}{*}{Model}    & \multicolumn{2}{c}{Fake}    &\multicolumn{2}{c}{Real} &\multirow{2}{*}{F1-score}               \\   
  & {Prec.} & {Rec.} & {Prec.}  & {Rec.}    &          \\  
  \midrule
  BERT & 0.69 &  \underline{0.74} & 0.73    & 0.69  &  \underline{0.71} \\   
  Bi-LSTM & 0.63 & 0.73 & \underline{0.80}   & 0.53    & 0.67             \\
  MDFEND & 0.66 & 0.59 & 0.38   & \textbf{0.92}    &  0.63            \\
  Pref-FEND & 0.71 & 0.47 & 0.63 & 0.87 & 0.65 \\   
Text+Relation & \underline{0.73} & 0.61 & 0.69 & 0.77 & 0.70 \\
  SemCAFE & \textbf{0.79}  &  \textbf{0.83}  & \textbf{0.82} & \underline{0.89}  & \textbf{0.83}                \\     \hline 
 \end{tabular}
 \label{tab:result1}
 \end{table}

\begin{table}[H]
\centering
\caption{Result of different classification models for news articles on Micro F1 score}
    \begin{tabular}{cccccc}  
    \toprule
    \multirow{2}{*}{Model}    & \multicolumn{2}{c}{Fake}    &\multicolumn{2}{c}{Real} &\multirow{2}{*}{F1-score}               \\   
  & {Prec.} & {Rec.} & {Prec.}  & {Rec.}    &          \\  
  \midrule
  BERT  & \underline{0.76}    & \underline{0.59}  & 0.63 &  \underline{0.83} &  0.66 \\   
  Bi-LSTM & 0.65   & 0.55   &  0.58 & 0.68  & 0.60             \\
  MDFEND & 0.69 & 0.54 & \underline{0.75}   & \textbf{0.86}    &  \underline{0.72}            \\
  Pref-FEND & 0.72 & 0.36 & 0.54 & 0.91  & 0.55 \\   
Text+Relation & 0.44 & 0.45 & 0.73 & 0.72 & 0.70 \\
  SemCAFE    & \textbf{0.78} & \textbf{0.66}  & \textbf{0.84}  &  \underline{0.83} & \textbf{0.78}               \\     \hline 
 \end{tabular}
 \label{tab:result2}
 \end{table}

\vspace{-2mm}
\subsubsection{Analysis of Result Obtained} 
\label{sec:result}
Results obtained from the classification model are reported in Table~\ref{tab:result1} and ~\ref{tab:result2}. We present both classes' precision, recall, and macro and micro F1 scores. We highlight the best score in bold and underlined the runner-up score. The result obtained from different baseline methods, including semantic relation, shows that our proposed method, in terms of macro score, SemCAFE, outperforms others; the closest match is the BERT model with different feature combinations such as content as text, title as text \cite{bracsoveanu2021integrating}. We found that adding type classification increases classification performance. However, due to the lack of recent entities mentioned in the YAGO KB, for instance, \textit{Volodymyr Zelenskyy}, our model performance fluctuates in the range of 83\% - 86\%. At a more granular level, SemCAFE gives the best precision and recall for fake class while MDFEND gives the best recall for real and SemCAFE, the best precision for real class. SemCAFE provides better precision and recall for fake classes because unreliable news articles contain repetitive and user-attractive texts to catch users' focus and repurposed claims to spread fake news. However, SemCAFE gives better results than real classes because reliable news articles show a consistent writing process and more named entities such as countries and politicians. MDFEND is able to capture multi-domain fake news, such as fake news related to politics and international relations, so it gives better recall. However, a similar trend is observed when looking at the micro scores, where SemCAFE outperforms other models. The closest match was MDFEND due to its ability to capture fake news on multiple topics.
BERT also performed well in recognizing fake news articles. Micro-average prioritizes individual instances, making it sensitive to a few high-performing examples of news articles, especially in dominant classes. In contrast, macro-averaging treats all classes equally, offering a more balanced evaluation in scenarios with class imbalance. However, there is an improvement in performance in the macro average, while micro scoring gives better recall for reliable news articles.
Apart from that, it can also be seen that the SemCAFE approach outperforms other models, so semantic fingerprinting helps with the reliability of news articles. 


\section{Conclusions \& Future Work}
\label{sec:6}
In this study we introduced SemCAFE, a novel approach to assess web source reliability through entity-level analytics. In particular, SemCAFE derives semantics from entity-level analytics by exploiting type relationships inherent in the underlying KG. Apart from boilerplate removal and standard NLP processing, we derive the inherent semantics of web contents through semantic fingerprinting. By doing so, we can disclose the dependencies of conceptually similar but non-identical entity types of a KG. 
The results show that SemCAFE can be used to identify and flag content originating from unreliable web sources. Our comprehensive studies show the superiority of SemCAFE over state-of-the-art approaches for precision in minor, but recall there is a huge difference; overall, the F1 score improves by 12\% considering the macro average. The viability of our approach to real-life data collected across a multitude of web domains obtained from various countries/languages. One limitation of this approach is that SemCAFE was applied to historical data, which may introduce delays in the collection and extraction of semantic information. Implementing this method in real-time would necessitate robust computational resources to ensure efficient processing. 

Our work has some limitations; first, in terms of datasets, we collected reliable news articles from news aggregators, which might not include news articles published from Ukraine and Russia, so datasets do not cover the fake news spread in disputed countries. We collected unreliable news articles, which the fact-checking organizations debunk, and there might be several unreliable news articles that are not debunked and still circulating on the web. So, a comprehensive real-time dataset might better exploit our approach. Second, for semantic fingerprinting, we used YAGO KB (version 3.1, which was compiled in 2017) to extract WordNet categories, so the latest named entity might not included in the proposed approach, such as names of new politicians (e.g., Volodymyr Zelenskyy).

In future work, we aim to enhance our experiments beyond ``classical'' online news websites. For this purpose; we also aim to extend the studies to social media. Thus, we will improve the robustness of the data extraction pipeline when dealing with more ``noisy'' social media content (including a high number of typos, hashtags, acronyms, and out-of-KB named entities). Along these lines and in the context of explainable AI, we would like to extend our approach beyond binary assessments by providing explanations/rationales explaining why certain decisions have been undertaken. Last but not least, we would like to integrate SemCAFE into a comprehensive web source reliability assessment framework. To this end, we want to intertwine SemCAFE with existing open-source tools in the domain of fake news detection.

\section*{Data \&  Ethical Considerations}

The dataset collected in the present study follows the rules for strict data access, storage, and auditing procedures for the sake of accountability. The SemCAFE dataset, together with the ML/DL models generated in the study, will be available to the community upon acceptance of the study. The sample dataset is shared on GitHub\footnote{\href{https://github.com/Gautamshahi/SemCAFE}{https://github.com/Gautamshahi/SemCAFE}}.

\section*{Positionality}
Finally, the authors acknowledge their own positional bias in the analysis of the content, coming from a Western-centric perspective. We encourage researchers from diverse backgrounds to contribute to this and wider study of fake news.

\begin{acks}
This work was funded by the PROCOPE-MOBILITÄT 2023 for visiting research at the University of Caen, France, supported by the Science and Technology Department of the French Embassy in Germany.
\end{acks}

\balance
\bibliographystyle{ACM-Reference-Format}
\bibliography{unreliable-news}

\appendix
\section{Appendix}
\label{sec:appendix}
In this section, we provided a list of search keywords and some examples of reliable and unreliable news sources used in the study.

\subsection{Search String from Wikinews}
\label{sec:a2}

The list of search keywords extracted from the Wikinews categories is as follows:
\emph{russian invasion of ukraine, russian invasion, invasion of ukraine, russo-ukrainian war, ukrainian war, controversies, international relations, russia, ukraine, europe, military history, military, military history of russia, history of russia, conflicts, invasions by russia, invasions russia, invasions, invasions of ukraine, opposition to nato, nato, conflicts in territory of the former soviet union, soviet union, former soviet union, russian irredentism, irredentism, russian–ukrainian wars, belarus–nato relations, belarus, belarus–ukraine relations, russia–nato relations, ukraine–nato relations, russia–nato, russia nato, ukraine nato, ukraine–nato, vladimir putin, putin, vladimir, volodymyr zelenskyy, volodymyr, zelenskyy, alexander lukashenko, alexander, lukashenko, wars involving belarus, wars involving Chechnya, wars involving russia, wars involving the donetsk people republic, wars involving the luhansk people republic, wars involving ukraine, donetsk, luhansk
}

\subsection{List of Data Sources}
\label{sec:reliable} 
We present the list of news article's aggregators for reliable and unreliable articles as in Section~\ref{sec:dc} is shown in Table \ref{datasource}. We include Snopes (Fact-checking organization) because it is not covered by other news aggregators and represented by * in Table \ref{datasource}. The overall list of each news website for reliable news articles and fact-checking organization for unreliable news articles are discussed in Section \ref{sec:reliable} and \ref{sec:unreliable}.

\begin{table}[!htbp]
	\centering
	\caption{List of News Articles Aggregators}
\begin{tabular}{p{2.5cm} p{4.9cm}} 
\hline
\textbf{Source} & \textbf{Website}  \\ \hline

 & Reliable \\ \hline
Media Cloud & \url{https://www.mediacloud.org/}  \\ 
Newsdata.io  &  \url{https://newsdata.io/} \\ \hline

 & Unreliable \\ \hline
GFC    & \url{https://toolbox.google.com/factcheck/} \\
Pubmedia   & \url{https://fact.pubmedia.us/}  \\ 
EDMO   & \url{https://edmo.eu/}  \\ 
\#UkraineFacts  & \url{https://ukrainefacts.org/}  \\ 
Snopes (*)  & \url{https://www.snopes.com/} \\\hline

	\end{tabular}%
	\label{datasource}%
\end{table}

\subsection{List of Reliable News Sources}
\label{sec:reliable} 
We present some examples of reliable news sources and countries of origin; 261 sources will be shared along with data used in the study as mentioned in Section~\ref{sec:dc}. 
\begin{table}[H]
	\centering
	\caption{List of reliable news sources}
\begin{tabular}{p{2.3 cm} p{3.6cm}p{2.2cm}} 
\hline
\textbf{Source} & \textbf{Website}  & \textbf{Country origin}\\ \hline
New York Post  & \url{www.nypost.com}  & USA \\ 
India Today  &  \url{www.indiatoday.in} & India\\ 
The Japan Times   & \url{www.japantimes.co.jp} & Japan \\ 
Reuters & \url{www.reuters.com} & United Kingdom \\ 
Independent & \url{www.independent.co.uk}  & United Kingdon\\ 
Fox News  & \url{www.foxnews.com} & USA\\ 
Malay Mail   & \url{www.malaymail.com} & Malaysia \\ 
Ctv News  & \url{www.ctvnews.ca}  & Canada\\ 
The Conversation  & \url{www.theconversation.com}  & Australia\\ 
Forbes  & \url{www.forbes.com} & USA \\ \hline

	\end{tabular}%
	\label{reliablesource}%
\end{table}

\subsection{List of Unreliable News Sources}
\label{sec:unreliable}
We present some examples of unreliable news sources and countries of origin; a total of 581 sources will be shared along with data used in the study, as mentioned in Section~\ref{sec:dc}. 
\begin{table}[!htbp]
	\centering
	\caption{List of unreliable news sources}
\begin{tabular}{p{2 cm} p{3.2cm}p{2.2cm}} 
\hline
\textbf{Source} & \textbf{Website} & \textbf{Country origin}\\ \hline
Avavot  & \url{www.aravot.am}  & Armenia \\ 
Politnavigator  &  \url{www.politnavigator.net}  & Russia \\ 
Iz  & \url{iz.ru}   & Russia   \\ 
Tass & \url{tass.ru}  & Russia   \\ 
News-kiev & \url{news-kiev.ru}  & Russia \\ 
Lnr-news & \url{lnr-news.ru}  & Russia \\ 
Msdernet  & \url{www.msdernet.xyz} & USA \\ 
Hawamer  & \url{hawamer.com}  & Saudi Arabia \\ 
Masr  & \url{www.masr.today}  & Egypt \\ 
Sputnik  & \url{sputnik.by}  & Russia \\ \hline

	\end{tabular}%
	\label{data:rel}%
\end{table}

\end{document}